\documentclass[final]{cvpr}

\usepackage{times}
\usepackage{epsfig}
\usepackage{graphicx}
\usepackage{amsmath}
\usepackage{amssymb}
\usepackage{multirow}
\usepackage{xspace}
\usepackage{amssymb}
\usepackage{pifont}
\newcommand{\cmark}{\ding{51}}%
\newcommand{\xmark}{\ding{55}}%

\usepackage[pagebackref=true,breaklinks=true,colorlinks,bookmarks=false]{hyperref}

\def\cvprPaperID{3112} 
\def\confYear{CVPR 2021}
\begin{document}
\newcommand{\method}{{LTS}\xspace}

\title{


Locate then Segment: A Strong Pipeline for Referring Image Segmentation
}

\author{Ya Jing,\textsuperscript{\rm 1,2}\thanks{This work was done when Ya Jing was an intern at ByteDance AI Lab.}~~
Tao Kong,\textsuperscript{\rm 3}~~
Wei Wang,\textsuperscript{\rm 1,2}~~
Liang Wang,\textsuperscript{\rm 1,2}~~
Lei Li,\textsuperscript{\rm 3}~~
Tieniu Tan\textsuperscript{\rm 1,2}\\
\textsuperscript{\rm 1}Center for Research on Intelligent Perception and Computing (CRIPAC),\\
National Laboratory of Pattern Recognition (NLPR)\\
Institute of Automation, Chinese Academy of Sciences (CASIA)\\
\textsuperscript{\rm 2} School of
Artificial Intelligence, University of Chinese Academy of Sciences (UCAS)\\
\textsuperscript{\rm 3}ByteDance AI Lab\\
}

\maketitle
\pagestyle{empty}
\thispagestyle{empty}

\begin{abstract}
Referring image segmentation aims to segment the objects referred by a natural language expression.
Previous methods usually focus on designing
an implicit and recurrent feature interaction mechanism to fuse the visual-linguistic features to directly generate the final segmentation mask without explicitly modeling the localization information of the referent instances.
To tackle these problems, we view this task from another perspective by decoupling it into a ``Locate-Then-Segment" (\method) scheme.
Given a language expression, people generally first perform attention to the corresponding target image regions, then generate a fine segmentation mask about the object based on its context. 
The \method first extracts and fuses both visual and textual features to get a cross-modal representation, then applies a cross-model interaction 
on the visual-textual features to locate the referred object with position prior, and finally generates the segmentation result with a light-weight segmentation network. 
Our \method is simple but surprisingly effective. On three popular benchmark datasets, the \method outperforms all the previous state-of-the-arts methods by a large margin (\textit{e.g.,} +3.2\% on RefCOCO+ and +3.4\% on RefCOCOg). 
In addition, our model is more interpretable with explicitly locating the object, which is also proved by visualization experiments. 
We believe this framework is promising to serve as a strong baseline for referring image segmentation. 
\end{abstract}

\section{Introduction}
\label{sec:intro}
Jointly learning vision and language is a significant task in machine learning and pattern recognition community, which has drawn great attention in recent years. 
In this paper, we study the challenging task of language-instructed object segmentation \cite{hu2016segmentation,yu2018mattnet,ye2019cross} which aims to generate a segmentation mask of the object in image referred by a natural language expression. It has wide applications, e.g., interactive image editing and language-guided human-robot interaction. Beyond traditional semantic segmentation, language-referring image segmentation is more challenging due to the semantic gap between image and language. 
In addition, the textual expression is not just limited to entities (e.g., “person", “horse"). It may contain descriptive words, such as object properties (e.g., “red", “young"), actions (e.g., “standing”, “hold”), and positional relationships (e.g., “right”, “above”). 

\begin{figure}
\centering
\includegraphics[width=0.93\linewidth]{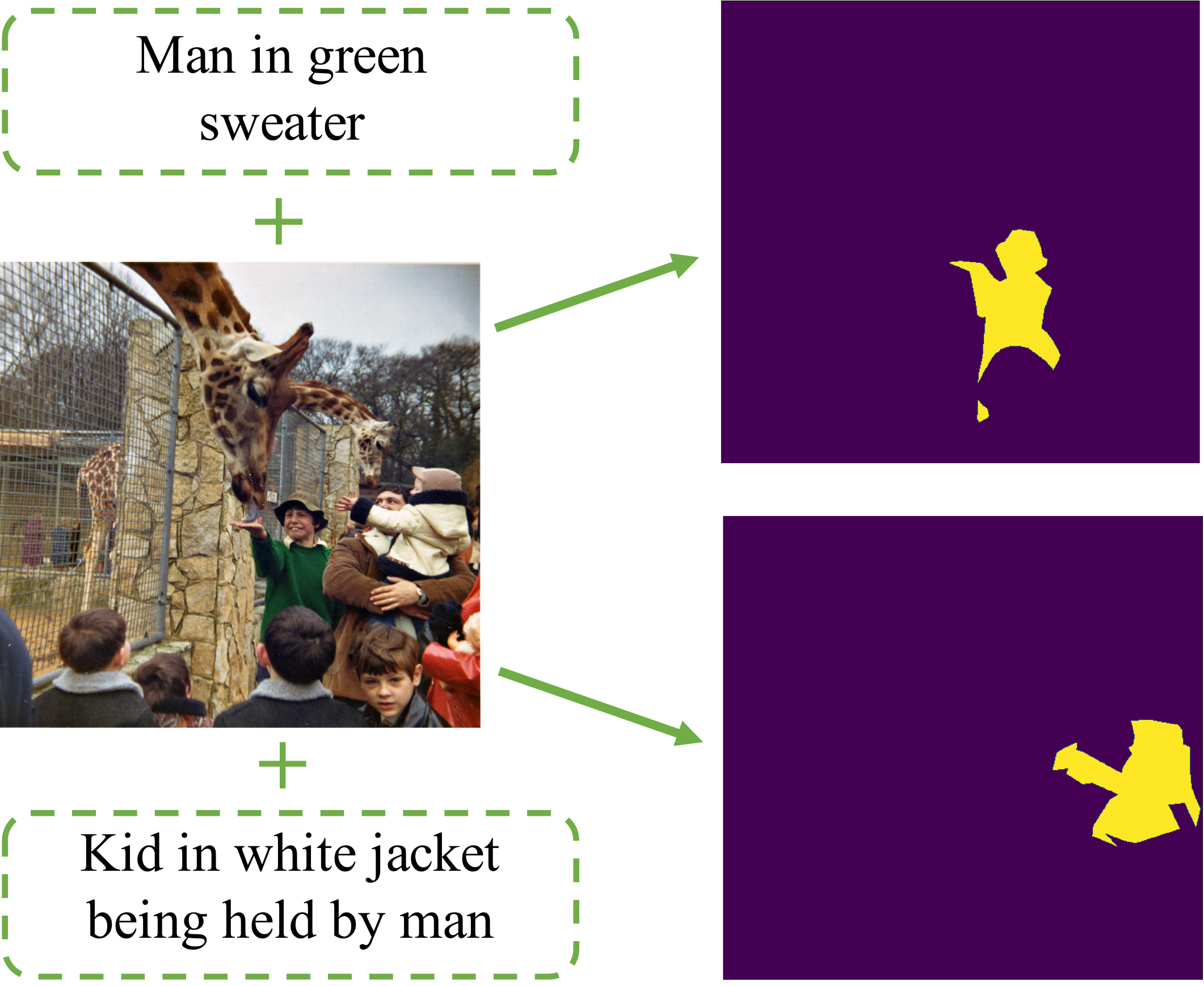}
\caption{The illustration of referring image segmentation. Given a referring expression and an image, the model aims to generate a segmentation mask of the corresponding object in image referred by the language expression. Best viewed in color.}
\label{fig:illstratue}
\end{figure} 


Given the image and referring sentence, 
there are two essential issues affecting the overall performance of a referring image segmentation model. 
First, the model must highlight the most discriminative candidate area in image corresponding to the given language.
Second, the model must generate a fine segmentation result.
The existing referring image segmentation methods could be generally summarized as follows: (1) Utilizing a Convolutional Neural Network (CNN) and a Recurrent Neural Network (RNN) to represent the image feature $f_v(I)$ and language feature $f_{text}(X)$, respectively. (2) Cross-modal attention and recurrent ConvLSTM are used to fuse $f_v(I)$ and $f_{text}(X)$ to get a coarse mask. (3) Dense CRF (DCRF) is further used as post-processing to get the final fine segmentation $M(I)$.


Previous works mainly focus on how to fuse the image feature and language feature.
A straightforward solution \cite{hu2016segmentation} is to utilize a concatenation-and-convolution method to fuse visual and linguistic representations to produce the final segmentation result. However, this method cannot model the alignment between image and language effectively due to the fact that the visual and textual information is modeled individually. 
To further model the context between multi-modal features, some prior methods \cite{shi2018key,chen2019see,ye2019cross} propose cross-modal attention by adaptively focusing on important regions in the image and informative keywords in the language expression. 
Recently, to exploit different types of informative keywords in the language and
learn the aligned multi-modal representations, some works \cite{huang2020referring,huilinguistic} either perceive all the entities that are referred by the expression or utilize the linguistic structure as guidance to segment the referent. 
Although great progress has been made, the network architecture and experimental practice have steadily become more and more complex. This makes the algorithm analysis and comparison more and more difficult. In addition, they do not explicitly locate the referred object guided by language expression and only utilize time-consuming post-processing DCRF to generate the final refined segmentation. 

In this paper, we consider solving this problem from  
another perspective.
We decouple the referring image segmentation task into two sub-sequential tasks: (a) referring object position prediction, 
and (b) object segmentation mask generation.
In our model, we first fuse the visual and linguistic features to get a cross-modal feature. Then
for (a), we propose a localization
module to directly obtain the visual contents prior corresponding to the expression. Such object prior will be used as a visual positional guidance for the subsequent segmentation module. For (b), we concatenate the object prior with the cross-modal features and utilize a light-weight ConvNets to get the final segmentation mask (Fig.~\ref{fig:main_model}). 

Our solution is very
simple but surprisingly effective. 
On three challenging benchmarks, i.e., RefCOCO \cite{kazemzadeh2014referitgame}, RefCOCO+ \cite{kazemzadeh2014referitgame} and RefCOCOg \cite{mao2016generation}, 
our model outperforms the state-of-the-arts methods by a large margin (\textit{e.g.,} +3.2\% on RefCOCO+ and +3.4\% on RefCOCOg). Extensive ablation studies also verify the effectiveness of each component of our method.






\section{Related Work}
\label{sec:related}
In this section, we briefly review the related work about prior studies on object segmentation, referring image localization and segmentation, and cross-model interaction. 

\subsection{Object Segmentation}
Object segmentation has achieved great advances in recent years based on Fully Convolutional Network (FCN) \cite{long2015fully}. FCN-based models transform fully connected layers in CNN into convolutional layers to train a segmentation model in an end-to-end way. 
DeepLab \cite{chen2017deeplab} replaces regular convolution with atrous (dilated) convolution to enlarge the receptive field of filters, leading to larger feature maps with richer semantic information. 
PSPNet \cite{zhao2017pyramid} proposes a pyramid pooling module to capture multi-scale information. 
Some other works \cite{badrinarayanan2017segnet,lin2017refinenet} exploit low level features containing detailed information to generate more accurate results. The instance segmentation area also achieved great progress based on Mask R-CNN~\cite{he2017mask} and FCNs~\cite{long2015fully,wang2020solo}.
In this paper, we study the more challenging segmentation problem whose semantic categories are referred by language expression. 

\subsection{Referring Localization and Segmentation}
Referring image localization aims to localize specific objects in an image referred by a language expression with a bounding box. 
Some works \cite{yang2019cross,hu2017modeling} model the relationships between image and language to obtain the most related objects.
MAttNet \cite{yu2018mattnet} decomposes the referring expression into subject, location and relationship to compute a more accurate matching score. 
The aim of referring image segmentation\cite{hu2016segmentation} is to localize the referred object with a segmentation mask rather than a bounding box. Hu et al. \cite{hu2016segmentation} utilize the concatenation of visual and linguistic features from CNN and Long Short-Term Memory network (LSTM) \cite{hochreiter1997long} to generate the segmentation mask. 
To obtain more accurate result, \cite{li2018referring} fuses multi-level visual features to refine the local details of segmentation mask. 
Multi-modal LSTM \cite{liu2017recurrent} is employed to sequentially fuse visual and linguistic features in multiple time steps. 
Dynamic filters \cite{margffoy2018dynamic} for each word further enhance multi-modal features. Shi et al. \cite{shi2018key} utilize word attention to model key-word-aware context. Recently, 
some works \cite{huang2020referring,huilinguistic} either perceive all the entities that are referred by the expression or utilize the linguistic structure as guidance to segment the referent. 
Multi-task collaborative network \cite{luo2020multi} achieves a joint learning of referring expression comprehension and segmentation. 
In this paper, we propose a localization module to locate the referred object with position prior and a segmentation module to obtain the final segmentation result.

\begin{figure*}
\centering
\includegraphics[width=0.98\linewidth]{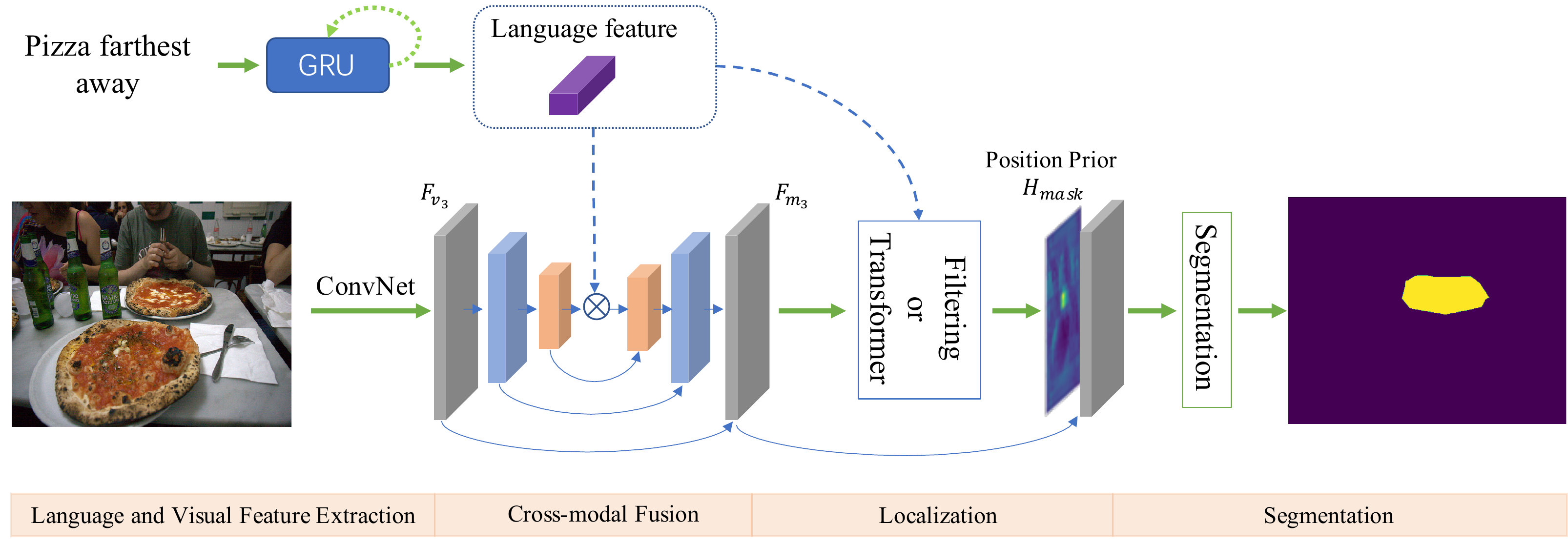}
\caption{The architecture of our proposed method.  The visual feature and linguistic feature are extracted by a deep convolutional network (ConvNet) and a bi-GRU network respectively, and then fused to generate the cross-modal features. Next a cross-modal interaction module (e.g., filtering and transformer \cite{vaswani2017attention}) is 
proposed to generate the object position prior. Finally, we concatenate the position prior and the cross-modal features to generate final segmentation mask by further convolutional refinement.}
\label{fig:main_model}
\end{figure*} 

\subsection{Cross-model Interaction}
In cross-modal tasks, a main challenge is to model the relationship between image and text. Recently, attention mechanism has been shown to be a powerful technique to extract the visual contents corresponding to the language expression in referring image segmentation. 

The relevance filtering can be seen as a simple way of attention mechanism, which is widely used in different areas of computer vision. Object tracking \cite{bertinetto2016fully} aims to localize an object in a video given the object region in the first frame, where relevance filtering is used to compare the first frame with the rest ones. Object classification \cite{wang2017residual} can be seen as a relevance filtering produce between output image feature and weight matrix of the last layer. 
Previously, relevance filtering has been considered in referring image segmentation \cite{margffoy2018dynamic}, but they use it implicitly to generate the final segmentation mask. 
In this paper, we utilize the direct language-conditional relevance filtering to obtain the relevance heatmap where higher response value is directly considered as the referred object prior. 

In addition to filtering, many cross-modal attention models \cite{shi2018key,chen2019see,ye2019cross} are proposed to adaptively focus on important regions in the image and informative keywords in the language expression. Different from them, we propose to utilize the unified attention-based building block transformer \cite{vaswani2017attention} to get the cross-modal relevance, which eliminates the need to design complex attention models.

\section{Proposed Approach}
\label{sec:method}
In this section, we explain the proposed \method in detail. First, we introduce the procedure of visual and textual representations extraction. Then we describe the two modules including filtering (or transformer) based localization and light-weight ConvNets based segmentation. Finally, we give the details of model learning process.

\subsection{Visual and Linguistic Feature Extraction}
As shown in Fig.~\ref{fig:main_model}, the input of our model consists of an image $I$ and a referring expression $X$. For similicity, we utilize ConvNets and GRUs to extract features of $I$ and $X$ respectively.

\textbf{Visual Feature} For the input image $I \in \mathbb{R}^{H \times W \times 3}$, we utilize the visual backbone to extract the multi-level visual features, which are denoted as $F_{v_{1}} \in \mathbb{R}^{\frac{H}{32} \times \frac{W}{32} \times d_{1}}$, $F_{v_{2}} \in \mathbb{R}^{\frac{H}{16} \times \frac{W}{16} \times d_{2}}$, and $F_{v_{3}}\in \mathbb{R}^{\frac{H}{8} \times \frac{W}{8} \times d_{3}}$, respectively. Note that $d$ is the dimension of feature channel, $H$ and $W$ are the height and width of the original image, respectively.

\textbf{Linguistic Feature} Given a referring sentence $X=[x_{1}, x_{2}, ..., x_{m}]$, where $x_{i}$ is the $i$-th token. We first apply table lookup to obtain the word embeddings. The embeddings are initialized as a 300-dimensional embedding vector by GLOVE embeddings \cite{pennington2014glove}. To model the dependencies between adjacent words, we use the standard bi-directional Gated Recurrent Unit (GRU) \cite{chung2014empirical} to handle the initial embedding textual vectors:
\begin{equation}
   \overrightarrow{h_{t}}  = \overrightarrow{GRU}(x_{t}, \overrightarrow{h_{t-1}}), h_{0} = 0,
\end{equation}
\begin{equation}
   \overleftarrow{h_{t}}  =  \overleftarrow{GRU}(x_{t}, \overleftarrow{h_{t+1}}), h_{m+1} = 0,
\end{equation}
where $\overrightarrow{GRU}$ and $\overleftarrow{GRU}$ represent the forward and backward GRUs, respectively. The global textual representation is obtained by average pooling between all word representations, which is donated as: 
\begin{equation}
   f_{text} = avg(h_{1}, h_{2}, ..., h_{m}),
\end{equation}
\begin{equation}
   h_{t} = concat(\overrightarrow{h_{t}}, \overleftarrow{h_{t}}), t \in [1,2,...,m],
\end{equation}

\textbf{Fusion} We obtain the multi-modal tensor by fusing $F_{v_{1}}$ with $f_{text}$, which is formulated as:
\begin{equation}
    f^{l}_{m_{1}} = g(f^{l}_{v_{1}} W_{v_{1}}) \cdot g(f_{text} W_{t}),
\end{equation}
where $g$ denotes Leaky ReLU, $f^{l}_{m_{1}}$ and $f^{l}_{v_{1}}$ are the feature vectors of $F_{m_{1}}$ and $F_{v_{1}}$, respectively, $W_{v_{1}}$ and $W_{t}$ are two transformation matrices to transform the visual and textual representations into the same feature dimension. Then, the multimodal tensors, $F_{m_{2}}$ and $F_{m_{3}}$ are obtained by:
\begin{equation}
 F^{'}_{m_{i-1}} = UpSample(F_{m_{i-1}})
\end{equation}
\begin{equation}
 F_{m_{i}} = concat(g(F^{'}_{m_{i-1}} W_{m_{i-1}}), g(F_{v_{i}} W_{v_{i}} )),
\end{equation}
where $i \in [2,3]$ and UpSampling has a stride of $2 \times 2$.
In the following process, we utilize $F_{m_{3}}$ as the input to generate the segmentation mask. Previous works usually adopt recurrent attention mechanism to get the segmentation results. In this paper, we show that locate-then-segment could get surprisingly superior performance, which will be introduces as follows.

\subsection{Localization}
To locate the object referred by language expression, we propose two ways to capture the context between multi-modal features including simple relevance filtering and unified attention-based block transformer, which eliminates the need to design complex attention model.

\textbf{Relevance Filtering}
The feature $F_{m_{3}}$ contains rich cross-modal information, which must be further modeled to get the relevant area in the image.
The aim of our cross-modality relevance filtering is to find the visual regions referred by the language expression, whose response scores are higher than the unrelated regions.
We first generate the language-guided kernel
 $K = f_{text} W_{k}$,
where $K \in \mathbb{R}^{d_{k}}$. Then it is reshaped into $\mathbb{R}^{d_{k} \times 1 \times 1}$ to perform filtering in fusion feature $F_{m_{3}}$:
\begin{equation}\label{eq:mask_prior}
 H_{mask} = conv(K, F_{m_{3}}),
\end{equation}
where $H_{mask} \in \mathbb{R}^{\frac{H}{8} \times \frac{W}{8}}$ and $conv$ means convolution operation. The heatmap $H_{mask}$ is a coarse segmentation mask where regions with higher response score means the more likely corresponding to the language expression (see Fig.~\ref{fig:main_model} position prior).

\textbf{Transformer}
To maintain consistency with the relevance filtering, here we do not utilize the transformer encoder to extract the textual representations but regard the global textual representation $f_{text}$ as the encoder output.

The decoder follows the standard architecture of the transformer, transforming the multimodal feature $F_{m_{3}}$ to response map $H_{mask}$ using multi-headed 
attention mechanisms:
\begin{equation}\label{eq:mask_prior2}
 {H}_{mask} = decoder(F_{m_{3}}, f_{text}).
\end{equation}

The decoder expects a sequence as input, hence we collapse the spatial dimensions of $F_{m_{3}}$ into one dimension, resulting in a $d\times\frac{HW}{64}$ feature map. Since the transformer architecture is permutation-invariant, we supplement it with fixed positional encodings \cite{bello2019attention} that are added to the input of each attention layer.

\subsection{Segmentation}
Given the visual object prior generated by Eq.~(\ref{eq:mask_prior}) or Eq.~(\ref{eq:mask_prior2}), the aim of the segmentation module is to generate the final fine segmentation mask.

\begin{figure}
\centering
\includegraphics[width=0.8\linewidth]{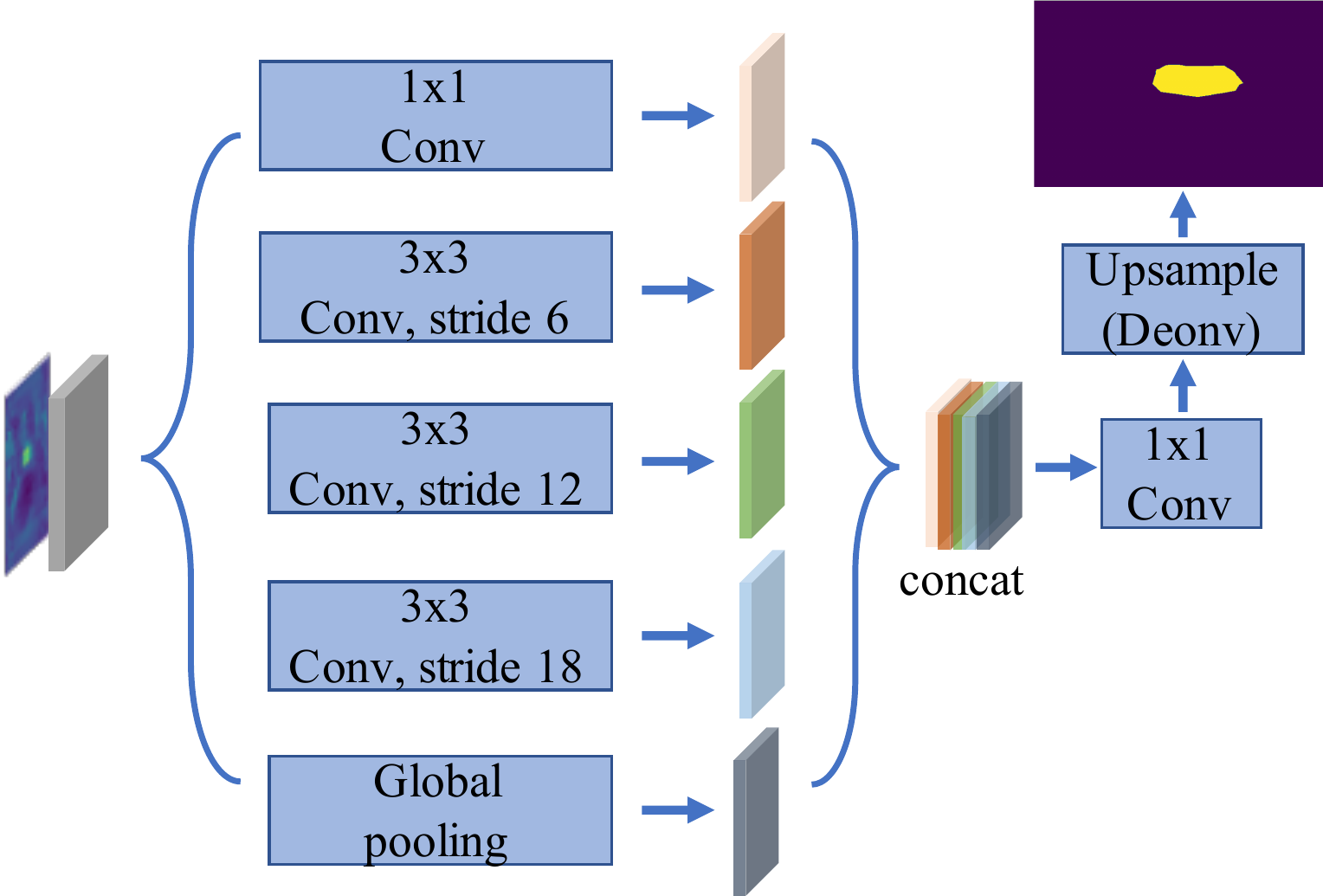}
\caption{The segmentation module. We first concatenate the feature $F_{m_3}$ and position prior map, and feed them into a single ASPP module, finally we upsample (deconvolution) the final generated mask.}
\label{fig:aspp}
\end{figure}

We first concatenate the original cross-modal feature $F_{m_3}$ and visual object prior $H_{mask}$, and utilize a segmentation module to refine the coarse segmentation result:
\begin{equation}
 P_{mask} = Seg(concat(F_{m_{3}}, H_{mask})),
\end{equation}
where the main structure of $Seg$ is ASPP \cite{chen2017deeplab}. The ASPP probes an incoming convolutional feature layer with filters at multiple sampling rates and effective fields-of-views, thus capturing objects as well as image context at multiple scales. Note that to obtain more precise segmentation results, we
adopt the deconvolution method to upsample the feature map
by a factor 2. Therefore, the predicted mask $P_{mask} \in \mathbb{R}^{\frac{H}{4} \times \frac{W}{4}}$. Fig.~\ref{fig:aspp} shows the segmentation process.

\subsection{Training and Inference}
During training, the Sigmoid Binary Cross Entropy (BCE) loss function is defined as follows:
\begin{equation}
 L_{seg} = \sum_{l=1}^{\frac{H}{4} \times \frac{W}{4}} [y_{l}\log(p_{l}) +(1-y_{l})\log(1-p_{l})],
\end{equation}
where $y_{l}$ and $p_{l}$ are the elements of the down-sampled ground-truth mask and predicted mask $P_{mask}$, respectively.

In addition, to make sure that the model can focus on the corresponding image regions, we add a locating loss to supervise the position prediction, which is defined as follows:
\begin{equation}
 L_{loc} = \sum_{l=1}^{\frac{H}{8} \times \frac{W}{8}} [y_{l}\log(h_{l}) +(1-y_{l})\log(1-h_{l})],
\end{equation}
where $h_{l}$ is the element of the down-sampled response map $H_{mask}$.

Finally, the total loss is defined as:
\begin{equation}
 L = L_{seg}+\lambda L_{loc},
\end{equation}
where $\lambda$ is empirically set to 0.1 in our experiments.

During inference, we upsample the predicted segmentation mask $P_{mask}$ to the original image size $H \times W$ and binarized at a threshold of 0.25 as the final result. No other post processing operations are needed.

\section{Experiments}
\label{sec:exp}
In this section, we first introduce the experimental setup including dataset, evaluation metrics, and implementation details. Then, we analyze the quantitative results of our method and a set of baseline variants. Finally, we visualize several segmentation masks.

\subsection{Experimental Setup}
\textbf{Datasets and Metrics}
We evaluate the proposed method on three benchmark datasets, i.e., RefCOCO \cite{kazemzadeh2014referitgame}, RefCOCO+ \cite{kazemzadeh2014referitgame} and RefCOCOg \cite{mao2016generation}.
We adopt intersection-over-union (IoU) and prec@X as the evaluation metrics \cite{luo2020multi,ye2019cross}. The IoU calculates intersection regions over union regions of the predicted segmentation mask and the ground truth. The prec@X measures the percentage of test images with an IoU score higher than the threshold $\gamma$ , where $\gamma \in \{0.5, 0.6, 0.7, 0.8, 0.9\}$.

The RefCOCO dataset contains 19,994 images with 142,210 referring expressions for 50,000 objects. The images and expressions are collected from the MSCOCO \cite{lin2014microsoft} with a two-player game \cite{kazemzadeh2014referitgame}. It is split into train, validation, test A and test B with a number of 120,624, 10,834, 5,657 and 5,095 samples, respectively. In general, each image contains two or more objects with the same object class and each expression has an average length of 3.5 words.

The RefCOCO+ dataset contains
19,992 images with 141,564 referring expressions for 49,856 objects. The images and expressions are also collected from the MSCOCO. It is also split into train, validation, test A and test B with a number of 120,191, 10,758, 5,726 and 4,889 samples, respectively. Different from RefCOCO, the expressions in RefCOCO+ include more appearances than absolute locations.

The RefCOCOg dataset is also collected from MSCOCO and contains 26,711 images with 104,560 referring expressions for 54,822 objects. Different from RefCOCO and RefCOCO+, expressions in RefCOCOg are collected from Amazon Mechanical Turk instead of a two-player game and have a longer length of 8.4 words includes both appearances and locations of the referent.
RefCOCOg \cite{mao2016generation,nagaraja2016modeling} have two types of data partitions, i.e., google partition \cite{mao2016generation} and UNC partition \cite{nagaraja2016modeling}. We adopt UNC partition in this paper.

\textbf{Implementation Details}
Following previous work \cite{luo2020multi}, we adopt the Darknet53 \cite{redmon2018yolov3} as the visual backbone, which is pre-trained on MSCOCO while removing the images appeared in the validation and test sets of three datasets. The input images are resized to 416$\times$416 and the input sentences are set with a maximum sentence length of 15 for RefCOCO and RefCOCO+, and 20 for RefCOCOg. A 1024 dimensional bi-GRU is used to extract the textual feature. The filtering dimension $d_{k}$ is set to 1024. The decoder has 1 layer network, 4 heads and 1024 hidden units. 
Adam \cite{Kingma2014Adam}
is used as the optimizer to train our model. The initial learning rate is 0.001, which is decreased by a factor of 0.1 at 30-th epoch. The batch size and training epochs are set to 18 and 45, respectively.

\subsection{Main Results}

\begin{table*}[htbp]
\begin{center}
\footnotesize
\caption{Comparison with the state-of-the-art methods on three benchmarks datasets using IoU as metric. ``-'' represents that the result is not provided. DCRF and ASNLS means DenseCRF \cite{krahenbuhl2011efficient} and ASNLS \cite{luo2020multi} post-processings, respectively.}
\label{table:sota}
\begin{tabular}{|l|c|c||c|c|c|c|c|c|c|c|}
\hline
\multirow{2}{*}{Methods}&\multirow{2}{*}{Backbone}&\multirow{2}{*}{DCRF}&\multicolumn{3}{c|}{RefCOCO}&\multicolumn{3}{c|}{RefCOCO+} &\multicolumn{2}{c|}{RefCOCOg} \\
\cline{4-11}
{}&{}&{}&{val}&{testA}&{testB}&{val}&{testA}&{testB}&{val}&{test} \\
\hline
\hline
{RMI \cite{liu2017recurrent}}&{ResNet101}&{\cmark}&{45.18}&{45.69}&{45.57}&{29.86}&{30.48}&{29.50}&{-}&{-} \\
\hline
{DMN \cite{margffoy2018dynamic}}&{ResNet101}&{\xmark}&{49.78}&{54.83}&{45.13}&{38.88}&{44.22}&{32.29}&{-}&{-} \\
\hline
{RRN \cite{li2018referring}}&{ResNet101}&{\cmark}&{55.33}&{57.26}&{53.95}&{39.75}&{42.15}&{36.11}&{-}&{-} \\
\hline
{MAttNet \cite{yu2018mattnet}}&{MRCN-ResNet101}&{\xmark}&{56.51}&{62.37}&{51.70}&{46.67}&{52.39}&{40.08}&{47.64}&{48.61} \\
\hline
{NMTree \cite{liu2019learning}}&{MRCN-ResNet101}&{\xmark}&{56.59}&{63.02}&{52.06}&{47.40}&{53.01}&{41.56}&{46.59}&{47.88} \\
\hline
{CMSA \cite{ye2019cross}}&{ResNet101}&{\cmark}&{58.32}&{60.61}&{55.09}&{43.76}&{47.60}&{37.89}&{-}&{-} \\
\hline 
{Lang2seg \cite{chen2019referring}}&{ResNet101}&{\xmark}&{58.90}&{61.77}&{53.81}&{-}&{-}&{-}&{46.37}&{46.95} \\
\hline
{BCAM \cite{hu2020bi}}&{ResNet101}&{\cmark}&{61.35}&{63.37}&{59.57}&{48.57}&{52.87}&{42.13}&{-}&{-} \\
\hline
{CMPC \cite{huang2020referring}}&{ResNet101}&{\cmark}&{61.36}&{64.53}&{59.64}&{49.56}&{53.44}&{43.23}&{-}&{-} \\
\hline
{MCN+ASNLS \cite{luo2020multi}}&{DarkNet53}&{\xmark}&{62.44}&{64.20}&{59.71}&{50.62}&{54.99}&{44.69}&{49.22}&{49.40} \\
\hline
{LSCM \cite{huilinguistic}}&{ResNet101}&{\cmark}&{61.47}&{64.99}&{59.55}&{49.34}&{53.12}&{43.50}&{-}&{-} \\
\hline 
{CGAN \cite{luo2020cascade}}&{DarkNet53}&{\xmark}&{64.86}&{\textbf{68.04}}&{62.07}&{51.03}&{55.51}&{44.06}&{51.01}&{51.69} \\
\hline
\hline
{\method (Ours)}&{DarkNet53}&{\xmark}&{\textbf{65.43}}&{67.76}&{\textbf{63.08}}&{\textbf{54.21}}&{\textbf{58.32}}&{\textbf{48.02}}&{\textbf{54.40}}&{\textbf{54.25}} \\
\hline
\end{tabular}
\end{center}
\end{table*}

To demonstrate the effectiveness of our model, we compare our segmentation results with the state-of-the-arts (SOTAs) methods on three referring segmentation benchmarks utilizing relevance filtering as the localization module, as shown in Tab.~\ref{table:sota}. It can be seen that our model achieves the best performances under IoU metric across different datasets even though we do not utilize the time-consuming post-processing, e.g., DenseCRF \cite{krahenbuhl2011efficient} and ASNLS \cite{luo2020multi}. Note that our model can further improve the performance when adopting relevance filtering for two times as shown in Tab.~\ref{table:filter}. Specifically, compared with the best competitor CGAN \cite{luo2020cascade} which proposes a cascade grouped attention network to perform step-wise reasoning, our model significantly outperforms it by about 3\% absolute IoU point on two challenging datasets (RefCOCO+ and RefCOCOg) performing only the simple relevance filtering once. The improved performances over the best competitor indicate that our model is very effective for this task.

\begin{figure*}[t]
\centering
\includegraphics[width=0.86\linewidth]{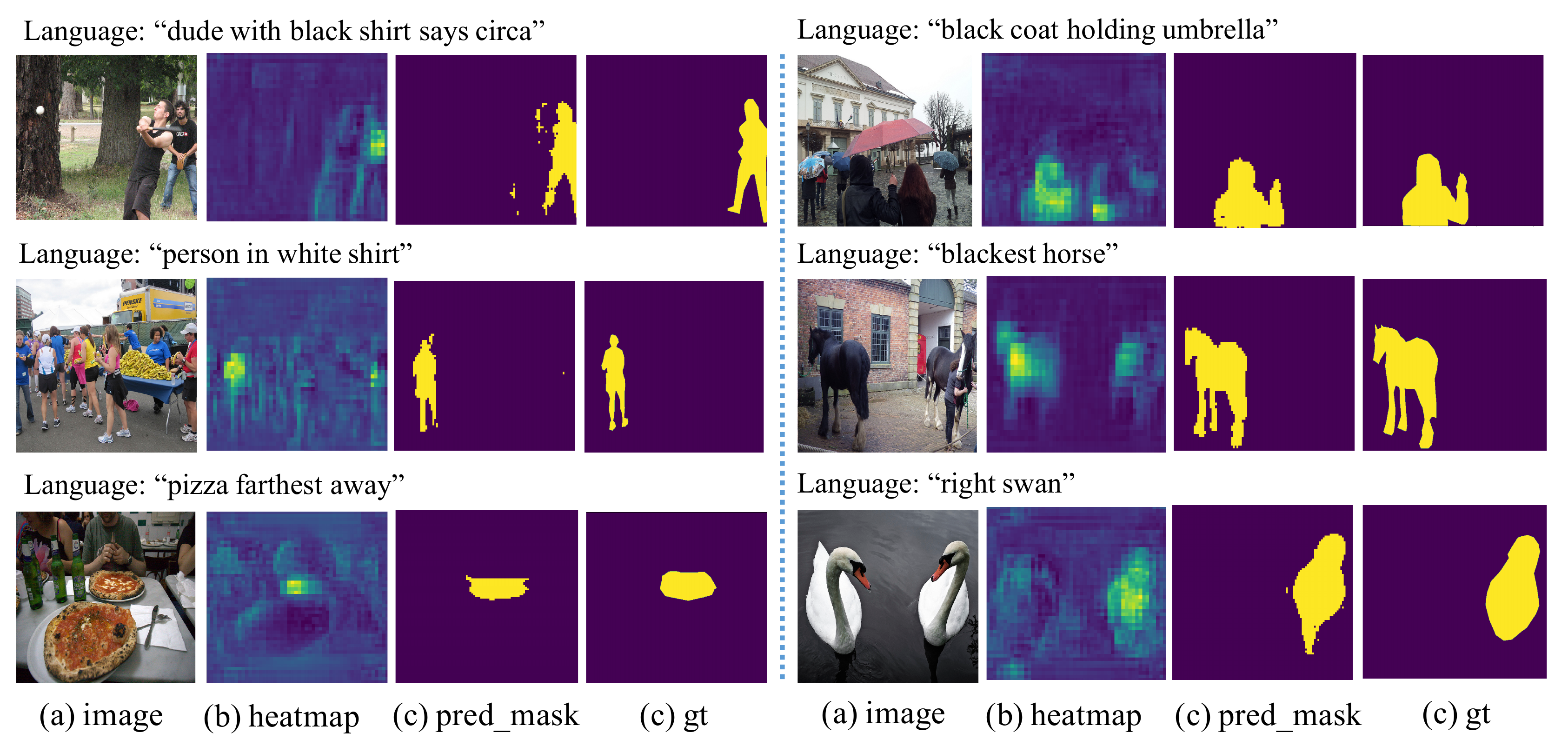}
\caption{Visualization of correlation heatmaps $H_{mask}$ (generated by relevance filtering) and final results $P_{mask}$ (pred\_mask) predicted by our model. gt means the ground truth segmentation mask of input image. Best viewed in color.}
\label{fig:filter}
\end{figure*}

Compared with the methods CMPC \cite{huang2020referring} and LSCM \cite{huilinguistic} which either perceive all the entities that are referred by the expression or utilize the linguistic structure as guidance to segment the referred object, our model achieves much better performances by explicitly
modeling the object position prior followed with a segmentation module, demonstrating the effectiveness of our pipeline. Here DMN \cite{margffoy2018dynamic} and Lang2seg \cite{chen2019referring} also model the multi-modal context by filtering.
But DMN utilizes every word to generate the kernel and performs convolution. Therefore, it needs to regress all the generated maps. Lang2seg utilizes the spatial-aware dynamic filters and needs to perform filtering in 7 times for different local regions. Different from them, our \method utilizes the sentence feature to generate one kernel and only needs filtering once. And the heatmap is utilized as the location to guide the generation of segmentation mask. 
The improved performance (16\% and 6\% IoU point) over them suggests that our segmentation module can learn a more accurate segmentation mask after obtaining the object location by language expression.



\subsection{Qualitative Results}

To verify whether the proposed localization module can obtain the correct location of the referent and how the segmentation module works, we visualize the response heatmap $H_{mask}$ (generated by relevance filtering) and segmentation results of several samples shown in Fig.~\ref{fig:filter}. We can see that our model is able to localize the referent by the language-dependent filtering. 
We also evaluate the localization quality (the probability that the maximum value of heatmap lies in the ground truth mask). The result is 81.74\%, which further demonstrates our localization capacity of the relevance filtering. However, this heatmap is far from an accurate segmentation mask. After refined by the segmentation module which captures objects and image context at multiple scales, we can obtain the final precise prediction mask.

\begin{figure*}[t]
\centering
\includegraphics[width=0.75\linewidth]{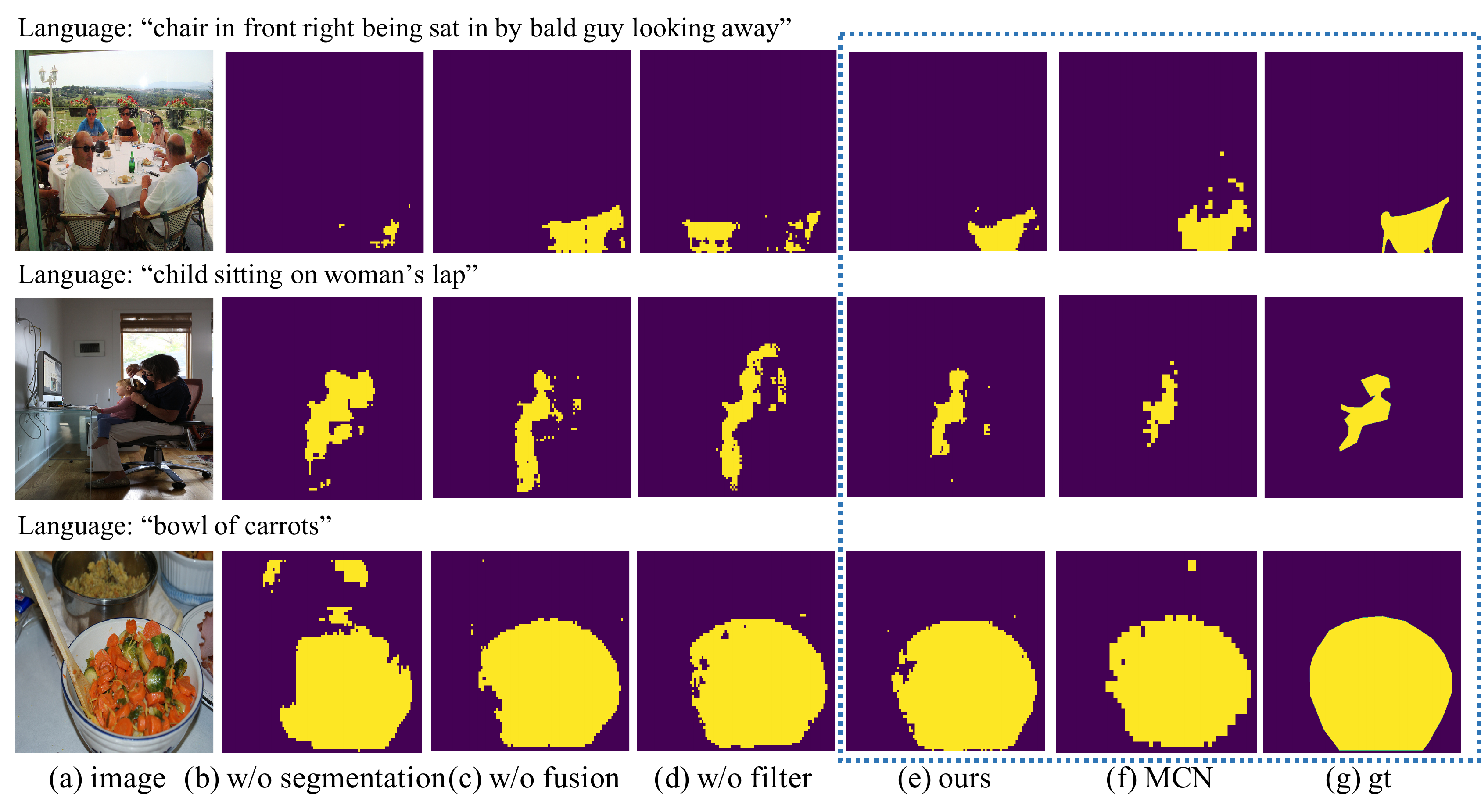}
\caption{Qualitative examples of referring image segmentation by different models. (b) (c) (d) show the proposed model w/o segmentation, fusion, and filter, respectively. MCN is the method proposed in \cite{luo2020multi}. Best viewed in color.}
\label{fig:visual}
\end{figure*}

\begin{table*}[t]
\begin{center}
\footnotesize
\caption{Ablation studies on RefCOCO dataset. Seg means the segmentation module.}
\label{table:abla}
\begin{tabular}{|c||cccc||c|c|c|c|c|c|}
\hline
{}&{Fusion}&{Filter}&{Seg}&{$L_{loc}$}&{IoU}&{prec@0.5}&{prec@0.6}&{prec@0.7}&{prec@0.8}&{prec@0.9} \\
\hline
\hline
{}&{\cmark}&{}&{}&{}&{55.08}&{60.58}&{51.26}&{41.55}&{27.15}&{6.76} \\
\cline{2-11}
{}&{}&{\cmark}&{}&{}&{54.53}&{60.43}&{50.00}&{39.03}&{24.95}&{5.85} \\
\cline{2-11}
{}&{\cmark}&{\cmark}&{}&{}&{56.94}&{63.49}&{54.20}&{44.13}&{29.43}&{8.36} \\
\cline{2-11}
{val}&{\cmark}&{}&{\cmark}&{}&{63.50}&{72.84}&{65.85}&{57.61}&{42.33}&{13.41} \\
\cline{2-11}
{}&{}&{\cmark}&{\cmark}&{}&{63.64}&{72.94}&{66.52}&{58.01}&{42.84}&{13.20} \\
\cline{2-11}
{}&{\cmark}&{\cmark}&{\cmark}&{}&{65.05}&{75.01}&{68.48}&{60.69}&{44.93}&{14.03} \\
\cline{2-11}
{}&{\cmark}&{\cmark}&{{\cmark}}&{{\cmark}}&{\textbf{65.43}}&{\textbf{75.16}}&{\textbf{69.51}}&{\textbf{60.74}}&{\textbf{45.17}}&{\textbf{14.41}} \\
\hline
\hline
{}&{\cmark}&{}&{}&{}&{56.82}&{62.49}&{53.60}&{42.87}&{28.27}&{6.65} \\
\cline{2-11}
{}&{}&{\cmark}&{}&{}&{56.05}&{61.66}&{51.76}&{40.32}&{25.61}&{5.67} \\
\cline{2-11}
{}&{\cmark}&{\cmark}&{}&{}&{58.53}&{64.52}&{55.65}&{45.59}&{30.51}&{7.94} \\
\cline{2-11}
{testA}&{\cmark}&{}&{\cmark}&{}&{65.31}&{75.32}&{69.45}&{60.46}&{44.76}&{11.38} \\
\cline{2-11}
{}&{}&{\cmark}&{\cmark}&{}&{66.41}&{77.00}&{71.27}&{62.63}&{46.65}&{12.82} \\
\cline{2-11}
{}&{\cmark}&{\cmark}&{\cmark}&{}&{67.49}&{78.10}&{72.87}&{64.47}&{\textbf{48.31}}&{\textbf{13.24}} \\
\cline{2-11}
{}&{\cmark}&{\cmark}&{{\cmark}}&{{\cmark}}&{\textbf{67.76}}&{\textbf{78.47}}&{\textbf{73.13}}&{\textbf{64.56}}&{47.98}&{12.92} \\
\hline
\hline
{}&{\cmark}&{}&{}&{}&{53.52}&{59.20}&{49.64}&{39.20}&{26.22}&{8.58} \\
\cline{2-11}
{}&{}&{\cmark}&{}&{}&{52.55}&{56.45}&{46.91}&{37.45}&{24.10}&{8.03} \\
\cline{2-11}
{}&{\cmark}&{\cmark}&{}&{}&{55.12}&{60.61}&{51.21}&{41.32}&{29.07}&{10.64} \\
\cline{2-11}
{testB}&{\cmark}&{}&{\cmark}&{}&{60.97}&{69.13}&{62.24}&{53.39}&{39.65}&{15.68} \\
\cline{2-11}
{}&{}&{\cmark}&{\cmark}&{}&{60.72}&{68.03}&{61.14}&{52.60}&{39.59}&{16.68} \\
\cline{2-11}
{}&{\cmark}&{\cmark}&{\cmark}&{}&{62.43}&{70.99}&{\textbf{64.65}}&{55.27}&{42.41}&{\textbf{17.48}} \\
\cline{2-11}
{}&{\cmark}&{\cmark}&{{\cmark}}&{{\cmark}}&{\textbf{63.08}}&{\textbf{71.82}}&{64.59}&{\textbf{55.74}}&{\textbf{42.79}}&{17.35} \\
\hline
\end{tabular}
\end{center}
\end{table*}

Fig.~\ref{fig:visual} shows the segmentation masks obtained by different models, which demonstrates the benefits of each module in our proposed method.
\begin{itemize}
    \item The predicted mask from model w/o segmentation can only obtain the location of the referent but not the fine mask of object, as shown in column (b);
    \item The model w/o fusion or filter also generates lower quality masks compared with our proposed full model as they fail to make clear judgement of the referred object, as shown in column (c) and (d).
\end{itemize}

Here we also compare our results with MCN~\cite{luo2020multi} in column (e) and (f).
Our generated segmentation masks have more obvious object shapes and finer outlines.

\subsection{Ablation Experiments}
Our proposed model is mainly composed of three modules, cross-modal fusion, localization (relevance filtering or transformer) and segmentation.
To investigate these three components and the proposed locating loss in our model, we perform a set of ablation studies on the Refcoco dataset. Tab.~\ref{table:abla} shows the result.

\begin{table}[!ht]
\begin{center}
\caption{Results of utilizing segmentation module on CMPC \cite{huang2020referring} on the RefCOCO dataset using IoU as metric.}
\label{table:refine}
\begin{tabular}{|c||c|c|c|}
\hline
{Model}&{val}&{testA}&{testB} \\
\hline
\hline
{CMPC}&{61.36}&{64.53}&{59.64}\\
\hline
{CMPC+Seg}&{\textbf{62.75}}&{\textbf{65.34}}&{\textbf{61.08}}\\

\hline
\end{tabular}
\end{center}
\end{table}

\begin{table}[!ht]
\begin{center}
\caption{Results of utilizing transformer instead of filtering on the val sets of three datasets using IoU as metric.}
\label{table:trans}
\begin{tabular}{|c||c|c|c|}
\hline
{Model}&{RefCOCO}&{RefCOCO+}&{RefCOCOg} \\
\hline
\hline
{LTS}&{65.43}&{54.21}&{54.40}\\
\hline
{LTS-Trans}&{\textbf{66.15}}&{\textbf{54.52}}&{\textbf{54.51}}\\
\hline
\end{tabular}
\end{center}
\end{table}

\begin{table}[!ht]
\begin{center}
\caption{Effects of filtering on RefCOCO dataset using IoU as metric. $n$ denotes number of filtering times. WordFilter means utilizing every word feature to generate kernel as DMN \cite{margffoy2018dynamic}.}
\label{table:filter}
\begin{tabular}{|c|l||c|c|c|}
\hline
{Model}&{n}&{val}&{testA}&{testB}\\
\hline
\hline
{}&{1}&{65.43}&{67.76}&{63.08}\\
\cline{2-5}
{\method}&{2}&{\textbf{66.04}}&{\textbf{68.68}}&{\textbf{63.27}}\\
\cline{2-5}
{}&{3}&{65.54}&{67.82}&{62.97}\\
\hline
{\method-WordFilter}&{1}&{64.92}&{67.31}&{62.50}\\
\hline
\end{tabular}
\end{center}
\end{table}

We first investigate the importance of fusing textual feature with visual feature to build multi-modal representations. It can be seen that the IoU accuracy on val dataset
drops 2.4\% (Fusion+Filter vs Filter) and 1.4\% (Fusion+Filter+Seg vs Filter+Seg). The fusion module proves the effectiveness of multi-modal representation in learning the semantic alignment between visual and linguistic modalities. Then we investigate the importance of relevance filtering by removing it from Fusion+Filter and Fusion+Filter+Seg. The IoU accuracy on val dataset drops 1.9\% and 1.5\%, respectively, which demonstrates that obtaining the location of the referent by the language description is beneficial to enhance the segmentation results. Comparing the result between Fusion (Filter / Fusion+Filter) and Fusion+Seg (Filter+Seg / Fusion+Filter+Seg), we can find that segmentation module can effectively improve the performances by obtaining a refined segmentation mask. In addition, we add this proposed segmentation module on CMPC \cite{huang2020referring} as shown in Tab.~\ref{table:refine}. The improved performances demonstrate that our module can generate more precise prediction mask. Finally, we can see that adding the locating loss also obtains better performance by supervising the alignments between image and text.

Tab.~\ref{table:trans} shows the results when utilizing transformer instead of filtering as the localization module. Using more complex attention model can further improve the performance by locating the referent better. 

\begin{table}[!ht]
\begin{center}
\caption{Results of \method with different input resolutions on the RefCOCO dataset using IoU as metric.}
\label{table:size}
\begin{tabular}{|c||c|c|c|c|}
\hline
{Model}&{resolution}&{val}&{testA}&{testB} \\
\hline
\hline
{}&{320$\times$320}&{63.01}&{65.40}&{60.78}\\
\cline{2-5}
{}&{352$\times$352}&{64.04}&{66.24}&{61.76} \\
\cline{2-5}
{\method}&{384$\times$384}&{64.45}&{67.02}&{62.47} \\
\cline{2-5}
{(n=1)}&{416$\times$416}&{65.43}&{67.76}&{63.08}\\
\cline{2-5}
{}&{448$\times$448}&{65.71}&{67.90}&{63.23} \\
\cline{2-5}
{}&{480$\times$480}&{\textbf{65.90}}&{\textbf{68.16}}&{\textbf{63.45}} \\
\hline
\end{tabular}
\end{center}
\end{table}

\begin{table}[!ht]
\begin{center}
\caption{Results of utilizing transformer, more filters and larger input resolution on the val sets of three datasets.}
\label{table:best}
\begin{tabular}{|c||c|c|c|}
\hline
{Model}&{RecCOCO}&{RecCOCO+}&{RecCOCOg} \\
\hline
\hline
{LTS}&{65.43}&{54.21}&{54.40}\\
\hline
{LTS*}&{\textbf{66.75}}&{\textbf{54.94}}&{\textbf{54.51}}\\
\hline
\end{tabular}
\end{center}
\end{table}

Tab.~\ref{table:filter} shows the experimental results when adopting multiple relevance filters, where $n=2$ means we utilize relevance filtering twice in our model. When $n=2$, our method gets better performance (+0.61 IoU). Such score is much better than previous published best result. For simplicity, all other experiments are performed with $n=1$. Besides, we conduct an experiment by utilizing every word to generate kernel as DMN \cite{margffoy2018dynamic}. The results are shown in Tab.~\ref{table:filter}, where they obtain comparative performances. Considering the simplicity, we adopt sentence-based filtering in this paper.

In addition, we find that larger input resolution will improve the performance by providing richer information as shown in Tab.~\ref{table:size}. In this paper, we set the image to $416\times416$ for fair comparison with previous methods.

Furthermore, we perform the experiments with the setting of using transformer, more filtering times, and larger input resolution on RefCOCO, RefCOCO+, and RefCOCOg datasets. The results are shown in Tab.~\ref{table:best}. We can see that our model obtains better performances with this setting.

\section{Conclusion}
\label{sec:conclusion}
Referring image segmentation is a challenging task since it not only needs to discover the most relevant area given a language query, but also generate accurate object segmentation masks. In this work, we have developed a simple yet effective method for this task. Our approach decouples this task into two sub-sequential tasks: referring object prior prediction and fine object segmentation mask generation. 
Through explicitly modeling the position prior, we get much higher segmentation performance compared with previous best results. Extensive ablation studies verify the effectiveness of each component of our method.

Although the IoUs of our method are much higher than previous works, the mask quality is far from ground-truth (Fig.~\ref{fig:visual}). We believe recent progress on image segmentation such as rendering~\cite{kirillov2020pointrend} could give better mask quality. Besides, we only utilized simple visual and linguistic feature extraction backbones. More complex network structures have the potential to further improve the performance.

\section{Acknowledgments}
\label{sec:ack}
This work is supported by National Natural Science Foundation of China (61976214, 61721004, 61633021).

{\small
\bibliographystyle{ieee_fullname}
\bibliography{papers}
}

\end{document}